# Modified swarm-based metaheuristics enhance Gradient Descent initialization performance: Application for EEG spatial filtering

M. Moattari[†]

Biomedical engineering Department, Amirkabir University of Technology (Tehran Polytechnic), Tehran, Tehran, Iran, moatary@aut.ac.ir

M.H. Moradi

Biomedical engineering Department, Amirkabir University of Technology (Tehran Polytechnic), Tehran, Tehran, Iran, mhmoradi@aut.ac.ir

R. Boostani

Computer Engineering Department, Shiraz University, Shiraz, Fars, Iran, boostani@shirazu.ac.ir

**ABSTRACT**

Gradient Descent (GD) approximators often fail in the solution space with multiple scales of convexities, i.e., in subspace learning and neural network scenarios. To handle that, one solution is to run GD multiple times from different randomized initial states and select the best solution over all experiments. However, this idea is proved impractical in plenty of cases. Even Swarm-based optimizers like Particle Swarm Optimization (PSO) or Imperialistic Competitive Algorithm (ICA), as commonly used GD initializers, have failed to find optimal solutions in some applications. In this paper, Swarm-based optimizers like ICA and PSO are modified by a new optimization framework to improve GD optimization performance. This improvement is for applications with high number of convex localities in multiple scales. Performance of the proposed method is analyzed in a nonlinear subspace filtering objective function over EEG data. The proposed metaheuristic outperforms commonly used baseline optimizers as GD initializers in both the EEG classification accuracy and EEG loss function fitness. The optimizers have been also compared to each other in some of CEC 2014 benchmark functions, where again our method outperforms other algorithms.

# 1 Introduction

In engineering applications, using approximation-based search methods like GD is appropriate as their search space is convex around operating points [24]. However, the combination of a locally convex optimizer with a metaheuristic can lead to better initial seeds and is worth improving. In this paper, the GD performance is improved by initializer that searches in multiple locality scales. The purpose of this work is to optimize differentiable objective functions with fractal behavior or layers of monotonic nonlinearities. Some examples are perceptron neural nets or linear subspace methods. Other important functions are those with recursive objective functions, e.g., the power series eigen-decomposition[1]. Most of these functions represent nests of hills that are scattered in multiple scales and are eventually multi-scale. They are multi-scale in the sense of having local optimas scattered in varieties of scales of function-values in terms of overshoot or smoothness of their tangent per locality. To find a suitable metaheuristic as GD initializer, we have to consider algorithm complexity more seriously. Highly complex optimizers are not suitable for engineering applications, adding to their computational burdens.

For algorithms that handle complexity-accuracy trade-off one can name Bat Search, which simulates echolocation behavior of bats [3], Ant Colony that mimics pheromone impact in communications [4], Gravitational Search which is inspired by the law of the gravity [5], and artificial immune system (AIS) algorithms [6] to imitate behavior of animal immune system. These methods are capable of intensifying in localities as time goes on. The mentioned algorithms as well as simplistic optimizers [1, 2], despite plausibility due to low run-time in large-scale engineering applications, did not act effectively in cooperation with GD in the literature. As this paper's case study is reduced to learning problems with numerous convex locals (like neural networks and nonlinear subspace models), an optimizer has to be sought that acts more powerfully in multi-scale search, being able to intensify in the first runtime half as much as the last one. This is intractable in mentioned optimizers because they fail to learn the best configuration in hypothesis space with a huge number of plausible solutions scattered in different scales. The idea of hierarchy in organizations, which grants better seeds for stochastic approximation algorithms, yet has not been implemented on algorithms like PSO and ICA. Authors deliberately focus on PSO and ICA due to their simplistic mechanism and applicability in ranges of machine learning projects. Other state of the art metaheuristics either have very high time complexity, or they demand excessive memory. Therefore, other metaheuristics are not yet combined with GD in the literature [25].

---
[1] Please refer to Formula 11

To better initialize GD in the aforementioned engineering problems, the authors add multiple nests of swarms to PSO and ICA in a hierarchical fashion, and call such framework Organized Hierarchical Meta-heuristic (OHM). Organized hierarchical trends are evident in human behavior and cosmic systems. Organized hierarchical systems like civilized society and modular nests of systems in organisms are other examples for the considered systems. There is an organized hierarchy in the cosmic system which each component not only has a purpose for itself but also a goal for that level of the hierarchy. In such systems, the action of the most specific part is governed under control of all levels of hierarchy. There are various phenomena in nature based on organized hierarchies. Phenomena like:

- Conscious process is affected by a hierarchy of knowledge registered in the brain. Every level of consciousness is controlled by its higher level [17].
- Hierarchy in bureaucratic organizations and social relationships gives less freedom to lower level of hierarchy while providing more flexibility for higher levels[18].
- Hierarchical partitions in society from family and kinsfolks to city, state, country, and continent manifest elegant interplay between cooperation and individual competitions.

Based on these concepts, OHM framework is proposed to extend swarm-based algorithms like PSO and ICA and to improve GD initialization by the resulting meta-heuristics. The higher the level of hierarchy based on which a solution is updated, the more holistic and explorative the process gets. Therefore optimizer can initialize more suited solutions for wider convex localities. On the other hands, by selecting lower levels of hierarchies, elites in more narrow space become a basis for solution update. By randomly selecting any level, the algorithm will handle local and global search simultaneously, hoping to seek better solutions.

The objective of this paper is to improve gradient-based initializers in the context of swarm-based methods. Main contributions are:

1. Develop hybrid optimizer of GD with the proposed framework as its initializer. This offers GD a randomizer that searches a wider span of multiple scales to choose better seeds for GD convex regions. In such scenario, the initializer randomizes new instances after training batches during the GD update process.

2. Compare the new optimizers' performance among four types of single-modal, multi-modal, perturbed single-modal, and perturbed multi-modal benchmark series versus baselines commonly used in GD-hybrid applications. This part suggests the function group that suits better for the proposed framework and confirms the necessity of choosing differentiable multi-scale objective functions.

3. Compare cooperative version of OHM (PSO based) with its competitive version (ICA based) to gain more insights about framework role on PSO and ICA.

In the next section, detailed explanations of PSO and ICA are explained. After proposing the main framework, PSO and ICA will be empowered by it and the resulted variants are explained in section 3. Section 4 evaluates the hybrid variant of OHM with GD on an EEG study. Then it ends by comparing the proposed methods with their related counterparts over benchmark functions.

## 2 Required Concepts

A more detailed explanation about the function of PSO and ICA is necessary. Not only because they have entailed approaches usable in the proposed framework; But also because they were inspired by consistent laws of nature, cooperation, and competition.

## 2.1 Particle swarm optimization

PSO is by default inspired by swarming behavior of particles in a search space of multiple dimensions. Swarms like fishes and birds have a sort of social behavior making them act and react in a common way and not deviate that much from normality. On the other hand, they have their own individualistic behavior. Kennedy and Eberhart have inspired this behavior and set up update formula (1) for unconstrained optimization problems [9,11]. The formula has two terms; the first is particle movement through the best particle's solution and the second one is its movement towards the best global solution so far found. The update formula and newly produced solution is the form below:

$$V_{id}(k + 1) = V_{id}(k) + \varphi_{1d}(P_{id}(k) − X_{id}(k))+\varphi_{2d}(G_d(k) − X_{id}(k)) \quad (1)$$

$$X_{id}(k + 1) = X_{id}(k) + V_{id}(k + 1) \quad (2)$$

where $V_i = [V_{i1}, V_{i2}, \ldots, V_{iN}]$ is the velocity for particle i; $X_i = [X_{i1}, X_{i2}, \ldots, X_{iN}]$ is the position of particle i; $\varphi_{1d}$ and $\varphi_{2d}$ are uniformly distributed random number and are generated independently for each dimension; $P_i = [P_{i1}, P_{i2},...P_{iN}]$ is the best position found by particle i; $G = [G_1, G_2,...G_N]$ is the best position found by the entire population; $N$ is the dimension of the search space and k is the iteration number. After each update, the possibility of the betterment of best current G and also best P of the updated particle has to be checked out so as to change those values if necessary.

Adding inertial weight to the PSO update formula is reported to motivate exploration and hinders algorithm from getting trapped in local optima [11]. By having a big inertial weight w(k) as shown in (2), the exploration rate increases. While in contrast, by setting that small, the algorithm returns to exploitation mode due to ignorance of subsequent update values.

$$V_{id}(k + 1) = V_{id}(k) +*\varphi_{1d}(P_{id}(k) − X_{id}(k)) +\varphi_{2d}(G_d(k) − X_{id}(k)) + w(k)V_{id}(k) \quad (3)$$

$$X_{id}(k + 1) = X_{id}(k) + V_{id}(k + 1) \quad (4)$$

The significant characteristic of PSO lies in simultaneously updating social and cognitive search component. Higher weights of social component lead algorithm to search locally and higher weights of cognitive component make algorithm do global search. The situation in which the particle initials are selected far from each other makes PSO ready for better local searching or global searching when necessary. In PSO, each hierarchy level has its own elite.

After proposing the main framework, it will be shown that PSO update formula can be resulted from setting the hierarchy level parameter to 2 with a special update type. The increasing level of the hierarchy and the same time using the cooperative approach of PSO in solution update has one advantage. When regions of particles do not overlap in search space, the algorithm can test out best hierarchy levels for considering their organizations as particles and even change it adaptively.

## 2.2 Imperialist competitive algorithm

Generally, ICA is taken from sociopolitical behavior of imperialistic competition. First of all, the algorithm defines multiple sets of solutions as countries. Like the behavior of real countries, algorithm regards some countries as imperialists and major remaining as colonies of them. The countries are assigned to the most powerful imperialists. Power of each imperialist is inversely related to best solution cost of the country. The failed countries in the competition, i.e. those having low power with respect to others, are condemned to removal. Powerful empires attract colonies of weak empires making them weaker [12].

The process of ICA starts from a segmented series of solutions. After initializing countries as groups of variables, either country is tagged as empires or as imperialists. Imperialist location can be set as the best solution among countries the empire entails. The assimilation is achieved by propelling colonies through imperialist. This movement is formulated in Eq. 5.

$$x_{new} = x_{old} + U(0,\beta) \times V_1 \quad (5)$$

Where U is a uniform random number generator with mean 0 and length β, $V_1$ is the direction towards imperialist and β is a custom variable.

Algorithm of general ICA is as below. While stopping condition is not met;
- Algorithm initialization: Random solutions generation. Empires initialization and their own colonies.
- Assimilation phase: Colonies are attracted toward imperialist states.
- Revolution phase: Random changes take place in countries solutions.

- Exchange phase: Replacement of better colony (with lower cost) with the existing empire.
- Imperialistic competition: Competition of imperialists to take possession of colonies of each other.
- Elimination of weakest empires.

While genetic algorithm models selfish genes in the survival of fittest, ICA alters GA's approach to survival of fittest societies. This makes colonies be identified as empires and the best of them tries to attract other members to itself and annihilate other empires at last.

## 3 The proposed framework and algorithms

While the particles in PSO are regarded as swarms, each empire in ICA can also be viewed as a swarm that assimilates imperialist. However, the way swarms interact in PSO with each other are cooperative compared to competitive nature of instances of swarms in ICA. As a result, an organized hierarchy framework can be imposed on both ICA and PSO. After an adequate number of updates in PSO, swarms lie in a pool of low-cost and high-cost solutions due to the fact that swarms will not be removed contrary to ICA case. That makes PSO have more tendency to explore unlikely localities w.r.t. ICA.

### 3.1 Organized hierarchical metaheuristic (OHM), the proposed framework

The proposed framework is established on two bases. First, search space can be scanned in different scales using multiple levels of hierarchy. Hierarchy in here is nested grouping of groups making it easier to deal with large and small regions simultaneously. Secondly, considering groups as organizations helps to define exclusive metrics for measuring the effectiveness of group members.

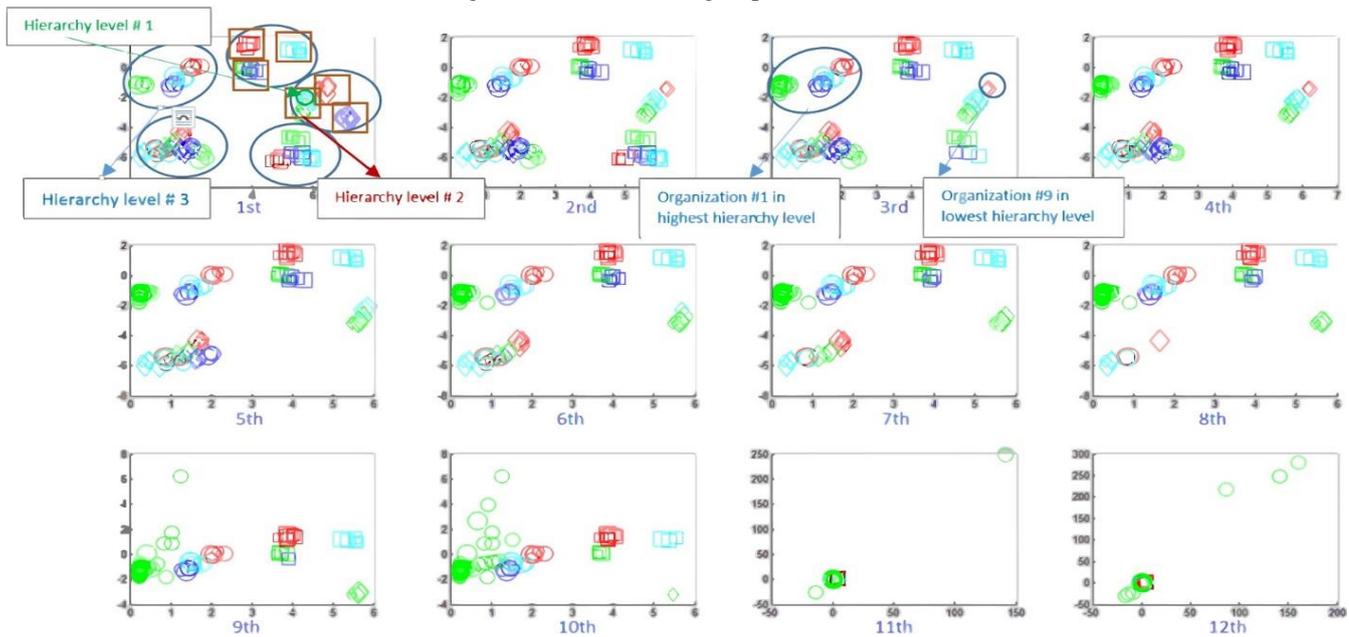

**Fig. 1: OHM update process of convergence and reduction of hierarchies.**

**Table 1: Solution selection types**

| Description | Tag | Description | Suitability | Process |
|---|---|---|---|---|
| Select elites (solutions with lower costs) | "FitnessRWS" | The lower the cost of solution in search space, the more likely it is to be updated | Search around more promising region (more intensifying) | - Convert costs to fitness by a subtracting from weight of least currently sought cost. - Use a roulette wheel search to select a solution |
| Select randomly | "UniformRWS" | All sought solutions have equal chance of selection | Leads to more exploration | - Define region of interest - Search randomly in the region |

**Table 2: Organization center Metrics. The location of search space in each organization region where the organization position gets identified with.**

| Metric | Metric Tag | Description |
|---|---|---|
| Location with minimum cost in the organization | "MinCostSolution" | The solution location with the least cost over all solutions of the organization is selected as the center of that organization |
| The weighted average of solution locations of organization in selected level entailing selected solution, with weights of solution fitness. | "WeightedMeanOfSolutions" | After selecting the level, the organization in that level is selected in which there exists a selected solution. All the containing solution locations get averaged by weights of their own fitness. Fitness is computed by subtraction of costs by a factor of lowest cost. |
| average of sub-organizations centers of organization in selected level entailing selected solution | "MeanCostSubOrganization" | Given the tag of an organization, Each of its sub-organization has its center recorded. By averaging their locations, the desired center will be acquired. |
| The weighted average of sub-organizations centers of organization in selected level entailing selected solution, with weights of each sub-organization fitness. | "WeightedMeanOfSubOrganizations" | Given the tag of an organization, Each of its sub-organization has its recorded center. By computing the weighted average of their locations with weights of their recorded organization selector (Table 3), the desired center will be acquired. |
| Center of organization ROI is selected as the organization center. | "RegionCenter" | In Fig. 2, the center of regions is shown per each hierarchy level. Region centers of the 1$^{st}$ level are shown by black stars. The deeper level has green stars as its centers. |

## Table 3: Selectors for selecting organization

| Selector Function | Selector Tag | Description | Suitability | Usage |
|---|---|---|---|---|
| Organization with maximum fitness in organizations of selected level | "MinCostSolution" | After selecting the hierarchy level, the organization with best solutions is selected among all organizations. | Organization with best solution can direct the starting solution to region with better promising solutions. | All |
| Roulete wheel selection of organizations based on their best soluion's fitness | "MinCostSolutionRWS" | After extracting least costs of solutions per each organization, they are converted to fitness by subtracting from factor of their minimum value. The fitness are used in a roultete wheel selector (RWS). RWS selects organizations with better solution in a more prioritized way. | Using roulete wheel can give previous metric more flexibility and lead to directing starting solution toward more promising regions. | All |
| Organization with least mean of costs | "MeanOfSubOrgCosts" | In the concerning hierarchy level, mean of costs for each organization are calculated. The organization with least calculated value is selected as metric output. | Using cost mean instead of minimum cost over interested region can lead to more realistic sense about regions fertility. | All |
| Organization with least mean of costs | "MeanOfSubOrgCostsRWS" | In the concerning hierarchy level, mean of costs for each organization are calculated. The organization with least calculated value is selected as metric output. | Using RWS in selecting organization can decreased biases in estimation of better regions. | All |
| Organization entailing the source solution | "EntailingOrg" | The solution selected in previous section (line) of algorithm has its own organization tag per level. The organization entailing that, is selected a metric output. | This metric permits the PSO enter OHM framework. By directing starting solution towards organization center it lies in, a more generalized PSO update rules is simulated. The higher the selected level of iteration, the more social the update takes place. Vice versa, the update resembles more to egoistic term in PSO update phase. | OHMPSO (PSO inspired) |
| Organization with least costs among the ones excluding source solution | "MinCostExcludingOrg" | Among all organizations in level of interest except which entails starting solution, least cost for each one is compared. Organization with least sought value will be returned as metric output. | This metric can be considered as a more general version of assimilation phase in ICA. The higher the level of hierarchy, the more explorative the imperialist's assimilation gets. In ICA the attractor empire was the most powerful of all. Similarly, in here the organization with least cost attracts the starting solution. | OHMICA (ICA inspired) |
| Organization with least costs among the ones excluding source solution | "MinCostExcludingOrgRWS" | Among all organizations in level of interest except which entails starting solution, least cost for each one is compared. Organization with least sought value will be returned as metric output. | Instead of strict selection of organization with best fitness, selection takes place with likelihood of their fitnesses. This gives the selection approach more flexibility. | OHMICA (ICA inspired) |
| Random selection of organization either using approach "MinCostExcludingOrg" or "EntailingOrg" | "EntailingOrgExcludingOrg" | Select randomly among organization which entails selected starting solutions and elite organization which excludes starting solution. Last two proposed approach of organization selection is combined using a random selection. If random number is less than a predefined value, the "EntailingOrg" approach is selected. Otherwise the "MinCostExcludingOrg" gets used. | This approach can control interplay between ICA update mode or PSO mode. The random value controlling selection likelihood of "MinCostExcludingOrg" or "EntailingOrg" approaches can be selected arbitrarily. | All |
| Random selection of organization either using approach "MinCostExcludingOrg" or "EntailingOrg" | "EntailingOrgExcludingOrgRWS" | Select randomly among organization which entails selected starting solutions and elite organization that excludes starting solution. Last two proposed approach of organization selection is combined using a random selection. If random number is less than a predefined value, the "EntailingOrg" approach is selected. Otherwise the "MinCostExcludingOrg" is used. | If metric MinCostExcludingOrg is selected, the organization is selected by RWS with lieklihoods of fitnesses. | OHHO (PSO ICA inspired) |

## Table 4: Parameter specifications and best-tested metrics for each proposed algorithm out of OHM

| Proposed algorithm | Associated metrics | | | | parameter preferences | | |
|---|---|---|---|---|---|---|---|
| Method | Sol. Sel. Mtr. Table 1 | Org. sel. Metric from Table 3 | Organization center from Metric Table 2 | Self-Tuning Mode. Tune rate of each hierarchy | Object Count In Each Organization | Hierarchy Count | Initial Level Effectiveness |
| OHMPSO | Fitness RWS | EntailingOrg | WeightedMean OfSolutions | No | [2 4 5 5] | 4 | [5 15 30 50] |
| OHMICA | Fitness RWS | MinCostExcludingOrg | WeightedMean OfSolutions | No | [2 4 5 5] | 4 | [5 15 30 50] |
| OHMPSO-ST (Self Tuning) | Fitness RWS | MinCostExcludingOrg | WeightedMean OfSolutions | Yes | [2 4 5 5] | 4 | [5 15 30 50] |
| OHMICA-ST (Self Tuning) | Fitness RWS | MinCostExcludingOrg RWS | WeightedMean OfSolutions | Yes | [2 4 5 5] | 4 | [5 15 30 50] |

## Table 5: Optimizers' preferences

| CLPSO | ICA | CICA | PSO |
|---|---|---|---|
| Inertia=[0.7,0.9] | Number of Countries: 60 | Number of Countries: 60 | Inertia= [0.7,0.9] |
| Speed limit: 0.1/8[Xmin, Xmax] | Imperialists=10 | Imperialists=10 | Speed limit: 0.1/8[Xmin, Xmax] |
| Possessed colonies | 120 | 120 | - |

### 3.1.1 Proposed framework's pseudo-code

To simulate ICA approach in the framework, the hierarchy level selector (i.e. organization) in Table 3 should be MinCostExcludingOrg, because for assimilating the elite, it excludes the swarms in which that solution resides. PSO selector, on the contrary, is set to EntailingOrg. It moves towards each of its swarms in elites containers that reside in the desired scale. For a detailed demonstration, please refer to Fig. 2. These selectors are remained fixed after initialization. The framework is summarized in the following algorithm:

---

ALGORITHM 1: Proposed framework's pseudo-code

---

Input: Objective function as cost.
Output: Best global solution. Framework process:
- Define the number of hierarchy levels, nested bounds of each hierarchy, organizations location, centers and maximum size of each level
- Randomly generate solutions covering search space
- Tag each solution to its covering hierarchy (Refer to the Initialization section of the Flowchart in Fig. 3 )
- Select a solution to update using Select metric I which suggested in Table 1
- Select hierarchy level based on the effectiveness of levels
- Select an organization in that hierarchy level based on metric II Table 3
- If rand < Update_Randomly_likelihood_Threshold
    o Randomly consider a set point location inbound of the selected organization
- else
    o Consider set point location as a selected organization center
- Move the solution toward the considered location
- Update center of a changed organization in all hierarchies by metric III suggested in Table 2
- Find the cost of the new solution
- Remove low costs solutions
- Modify or prune hierarchies
- Update effectiveness

The flowchart in Fig. 3 discusses the framework of OHM. First, the main parameters are defined to make algorithm ready for the initialization process. Main parameters for algorithm definition is listed in Table 8. After parameter definition, the hierarchy is initialized. Each hierarchy level entails a list of organizations with a specified center and bound. Depending on the number of hierarchy levels, organization centers are produced. The organizations in new hierarchy level are created in the neighborhood of each organization center of the previous level. This process is repeated until the total number of levels is reached. At that stage, instead of organizations, solutions are created in the neighborhood of organizations center of deepest level.

The flowchart for initialization of hierarchies is shown in the first part of Fig. 3. The organization tag of each solution in each level is saved as a tag array for that solution. Tag array is considered as the address of each solution just like the address of a house in the city. In each iteration, a solution is selected either using a fitness based distribution or by a random selector with uniform pdf. Afterwards, a level is selected among levels of hierarchy. In the next step, depending on the metric (in Table 3), one organization is chosen in the level and solution is directed through a point in its corresponding region. The point is either selected randomly or with metrics in Table 2. By getting the cost of the new solution, low-cost solutions will be removed, effectiveness values will be tuned and hierarchy modification process will be run.

### 3.2 Proposed algorithms for the framework

As mentioned in Section 2, there are two behaviors for swarm optimizer, each of which having its own impact over solution updates: i.e., cooperation and competition. In PSO, the interaction of swarms are cooperative and they tend to unite and approach the best solutions. Contrarily, in ICA, solutions during the update do not move through elites in their own group. Instead, they harness solutions in other locations. The first approach makes every swarm explore wider space when they are far from elite; while the second approach avoids exploration and tends to exploit the localities containing imperialist. In the following subsections, two algorithms (OHMPSO and OHMICA) are extracted from the proposed framework, each focusing on its unique aspect while holding the framework's commonality.

Fig. 1 demonstrates a schematic view of hierarchies, organizations and the population convergence in a 2D objective function every 10 iterations.

#### 3.2.1 Proposed organized hierarchical version of Particle Swarm Optimization (OHMPSO)

PSO is based on swarming behaviors of particles. During each iteration, the selected solution not only moves towards the current elite of the whole region but also toward its particle's elite. Social component in PSO update term resembles solution update in the first hierarchy of OHM, and cognitive update term in PSO resembles update in subhierarchy. By adding more levels of hierarchy, the proposed OHM framework is constructed; except for the fact that PSO regards only the cooperative aspect of a reduced OHM framework. The word cooperative in this context means that updates only take place inside organizations and the sub-organizations are attracted to each other.

OHMPSO constrains OHM framework to move the starting solution towards organization center it entails. Major organization selector for attaining OHMPSO is metric with tag EntailingOrg in Table 3. The ingredients of this selector are tuned for least-cost-error averaged over 20 benchmark functions each being run 10 times among other organization selection metrics. In subsequent iterations, different levels of the hierarchy are selected to update the initial state toward its covering organization centers in diverse gamut. Cooperation lies in the way each organization helps its entailing solutions toward more promising regions. Selected organization in level, entails the starting solution. If the solution is attracted by other excluding organizations, the update will get a sense of competition rather than cooperation. This form of competition among different non-overlapping organizations is explained in the next method as OHMICA. The first row of Table 3 describes metrics and parameter specification of OHMPSO.

Fig. 1 shows the progress of hierarchies, organizations and solution structure of OHMPSO during optimization of an objective function with two dimensions of inputs. Hierarchy levels organizations are shown in the first and third plot. Each new plot belongs to results after 10 new iterations. Plots are entitled with a number outlining the runtime iteration index. In the plots, each hierarchy level is provided with different color and each sub-hierarchy is shown with a different shape. The more the optimization process goes on, the less the organizations are removed, as search space is pruned by better solutions. It is evident in the last plot that all remaining organizations are scattered on a smaller regions.

**3.2.2 Proposed organized hierarchical version of Imperialistic Competitive Algorithm (OHMICA)**

Comparing ICA with PSO leads to two differences. First, instead of a social update phase of PSO, there exists imperialist competition attracting individuals towards other colonies. Second, there exists no cognitive update phase in ICA. Cognitive update term of PSO helps it get a global taste in solution updates. Proposed selectors in Table 3, MinCostExcludingOrg, makes OHMICA work in a competitive way, while selects a hierarchy level randomly for scale specifications. Afterwards, the starting point moves towards the center of organization which has the best solution. This approach simulates competition phase of ICA. Although the organization selector is fixed due to the structure of the algorithm, OHMICA can also get used with other organization center update metrics. Fig. 2 shows schematics of OHMPSO and OHMICA in parts (a) and (b) respectively. The blue star shown in each part is the selected solution for the update process. In part (a), solution moves towards the organization center of its own organization, leading to cooperation and intensification. Part (b) shows the solution moves towards centers of other hierarchy, giving a sense of competition to the algorithm, making it more explorative.

Most suitable parameters for the algorithm are selected and shown. Among all proposed selectors/metrics of Tables 1-3, parameter preferences have been chosen and shown in Table 4. Parameters performance are evaluated on average over absolute errors from 20 benchmark functions shown in Table 6, 7 in 10 runs.

# 4 Experimental Analysis and discussion

Three types of evaluations are approached to assess the capability of the proposed OHM framework. First and the main one is a hybrid version of Gradient Descent (GD) in which OHM is used as a random initializer. The objective function belongs to EEG subspace learning problem with sum of convex functions. As each convex region has its own width and scale, our approach is needed to search for initial seeds of GD more efficiently. The second evaluation finds global minima error of selected benchmark functions in a single setting with no approximators or hybrid modes. This test confirms the positive role of organized hierarchy in both competition and cooperation in two exploration and exploitation schemes. At last, the third evaluation compares different variants of OHM with each other to analyze the errors. The detailed descriptions of results are shown in the next three subsections.

Optimizer preferences are cleared up in Table 4. Proposed algorithms have been compared to set of algorithms which were either ICA based, or PSO based and also self-tuning-mode of them in which hierarchy level selector uses fitness distribution of currently sought solutions by a Roulette Wheel Selector to select hierarchy level.

In order to make OHM ready for global optimization of learning based models, OHM has to be implemented on simple ICA or PSO models. Equipping more complex variations of ICA or PSO with our proposed method, makes optimization time-consuming and intractable. As the objective of this study is to improve gradient-based initializers in context of swarm-based methods, a thorough fine-tuning has not been conducted on hyper-parameters. Swarm-based methods are easy to use, simple to implement, have high flexibility to modify and tune, and are the only algorithms hybridized with GD most frequently in the literature. Therefore, the authors preferred to use basic swarm-based context rather than baseline optimizers from CEC SOTA.

All the experiments are carried out in the same computer with Intel Core I 5 2.7GHz CPU, 3.00GB memory, and a Win7 32bit operation system. For both single and hybrid phases, the algorithm has been run 20 different times and the evaluation results of every run are recorded.

## 4.1 Optimization of subspace filtering objective function using hybrid OHM-GD method and comparison with other hybrid methods of GD

The hybrid optimizer in this case-study consists of two algorithms, GD, and its initializer as our proposed OHM method. It is evaluated on a subspace-learning problem named as weighted Common Spatial Patterns over trials. In order to define the main objective function, first, the reader has to get familiar with the Common Spatial Pattern (CSP).

**4.1.1 Common spatial pattern (CSP)**

CSP is a feature extraction method for classifying brain data in Brain Computer Interface. Due to its success in many neuroscientific case studies, it has turned into one of the most commonly used approaches for EEG features dimensionality reduction. The goal of CSP is to find projections of data to lower dimensions that have maximal data-variance per one label while having minimal variance per other labels [27]. For a single trial, an N-channel spatial-temporal EEG signal, let E be $N \times T$ matrix where T is the number of samples per channel. The normalized covariance matrix of the EEG is:

$$C = \frac{E^T E}{trace(E^T E)} \tag{6}$$

In the standard scheme, covariance matrices of each class (i.e. $C_1$ and $C_2$ in two-class settings), is computed by averaging over trials. The averaging process is not weighted. The objective is to find vector $w_k$ for $0 < k \in N < K_{components}$ such that:

$$w_k = \text{ArgMax}_{w\_k} \frac{w_k' C_i w_k}{w_k' (C_1+C_2) w_k} \tag{7}$$

Where each $w_k$ is orthonormal w.r.t. each other, and i is class label index, which here is either 1 or 2. As the Eq. (7) is a special case of Rayleigh quotient [28], it can turn into generalized eigenvalue problem by adding the denominator as Lagrangian to the numerator [29]. Differentiating w.r.t. $w_k$ leads to :

$$W \Lambda W' = (C_1 + C_2)^{-1} C_i (C_1 + C_2) \tag{8}$$

Where W is matrix out of $w_k$ as its columns and $\Lambda$ is the diagonal matrix of Lagrange multipliers in (7).

Eq. (8) is solved by eigenvalue decomposition methods like power iteration.

To evaluate the effect of the hybrid optimizer, similar methods to CSP like the ones in the next two sections have been brought about for validation accuracy baseline.

### 4.1.2 Devlaminck's work

The method proposed by Devlaminck et al. [31] assumes a similarity between spatial filters extracted from different subjects. The goal of this CSP variant is to construct a more global feature space by decomposing the spatial filter $w_i$ for each subject i into a global $w_0$ and subject-specific part $v_i$, i.e., $w_i = w_0 + v_i$,

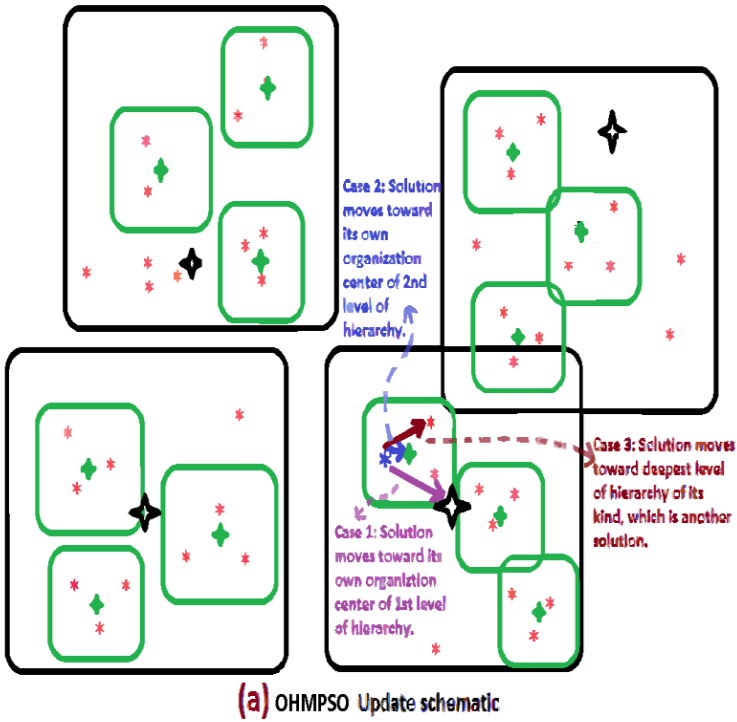

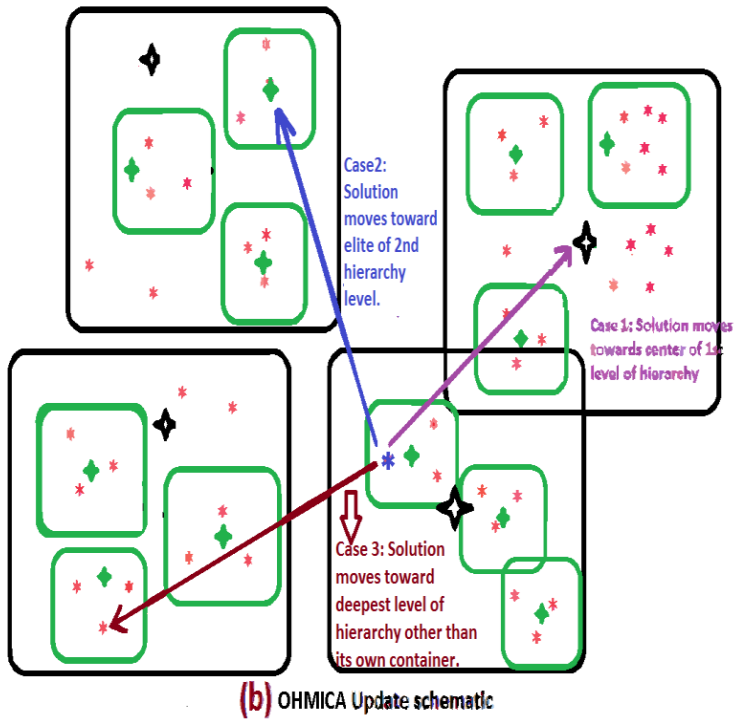

**Figure 2:** Schematics of OHMPSO and OHMICA algorithms for the solution update process. Star in each color, being elite among all solutions, resides in the same hierarchy level shown by the color frame.

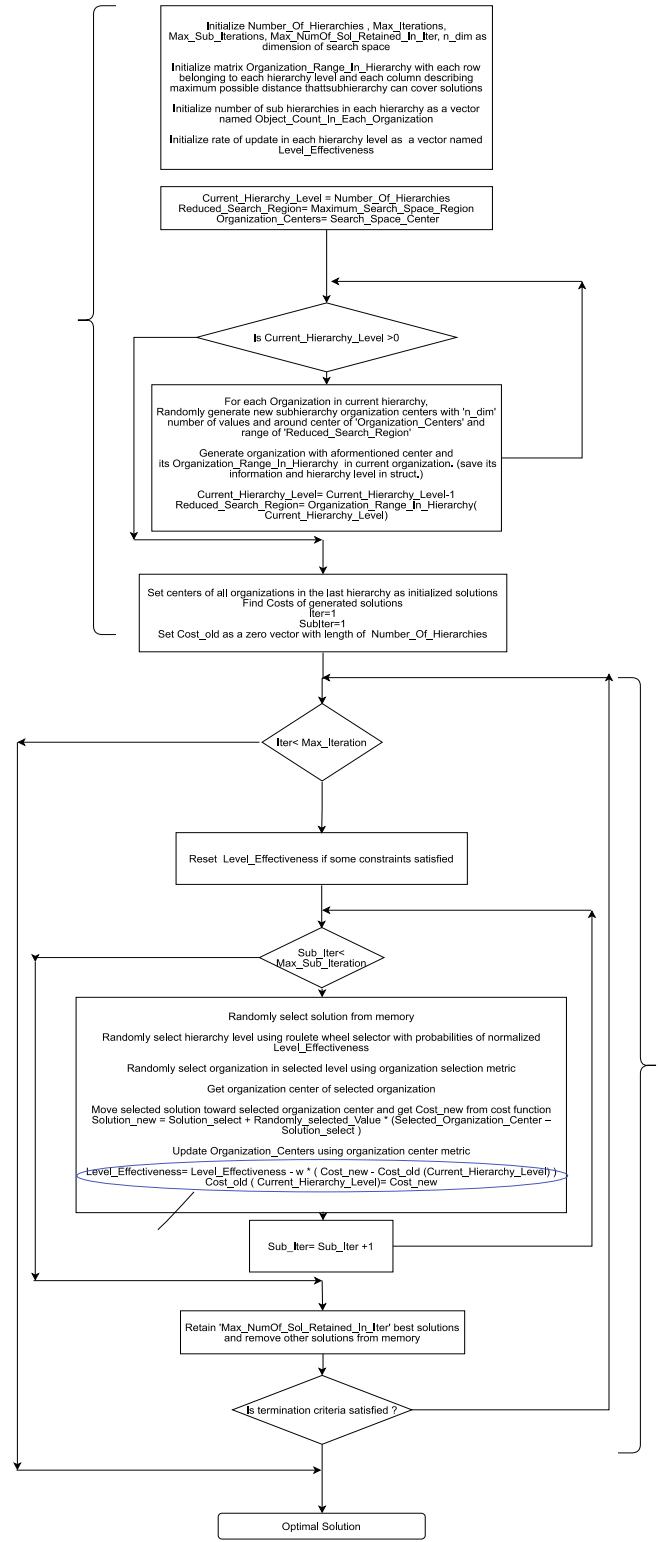

**Figure 3:** Detailed flowchart of OHM framework.

Applying a single optimization framework to learn both types of filters

$$\max_{a_{sk},w_k,v_{sk}} \sum_{k=1}^{K} \sum_{s=1}^{S} a_{sk} \frac{w_{sk}^T \Sigma_s^{(1)} w_{sk}}{s_{sk}^T \Sigma_s^{(2)} w_{sk} + \lambda_1 ||w_k||^2 + \lambda_2 ||v_{sk}||^2}, \quad a_{sk} \epsilon \{0,1\}, \quad \sum_{k=1}^{K} a_{sk} = 1 \qquad (9)$$

This method tries to optimize CSP filter by a weighted average of covariance matrices over trials for finding appropriate weights. However, due to binary weights, the weights cannot be approximated using gradient methods. Decomposing the filter to two parts of subject dependent and subject independent part makes this method a powerful baseline in the context of our proposed objective function in 4.1.3.

**4.1.3 Lotte and Guan's work**
The method proposed by Lotte and Guan regularizes the estimated covariance matrix by taking the mean covariance matrix of other subjects [32]. This kind of regularization may largely improve the estimation quality of high dimensional covariance matrix for scarce data. The estimation for subject i* can be written as

$$\tilde{\Sigma}_{i^*,c} = (1-\lambda)\Sigma_{i^*,c} + \frac{\lambda}{n-1}\sum_{i=1}^{n-1}\Sigma_{i^*,c} \qquad (10)$$

Where $\Sigma_{i*c}$ is the covariance matrix of class c for the subject of interest, $\Sigma_{i,c}$ are the covariance matrices of the other i = 1 . . . n, subjects and λ ∈ [0 1] is a regularization parameter controlling the amount of information incorporated from other users. This method is based on a restrictive assumption, namely the similarity between covariance matrices of different subjects.

This method is designed only for subjects and relationships between them. In contrast, the complexity of this model puts it in the same class with our proposed subspace model. As a result, it can be regarded as another suitable comparison baseline.

**4.1.4 The objective function for evaluation, Weighted Common Spatial Pattern (Wgt-CSP)**
To evaluate the proposed hybrid optimizer versus previous hybrids like GPSO, a new nonconvex objective function has been proposed. This new objective function is a modification to the CSP function (section 3.2.2.1), to not only improve the accuracy of speech imagery classification but also to assert the capability of the proposed hybrid optimizer OHM-GD over single GD and GPSO.

The objective is to maximize the following objective function in CSP framework:

$$\max_{b_i,a_i,w_0} \frac{w^T(\sum_{i=1}^{n} a_i \Sigma_{i,c}) w}{w^T(\sum_{i=1}^{n} b_i (\Sigma_{i,1} + \Sigma_{i,2})) w} \qquad (11)$$

Where $\Sigma_{i,c}$ is the selected epoch's covariance matrix of the class c which is NxN square matrix with N as the number of channels. w is one row of CSP projector. Each epoch could be ERP epoch. Depending of different $a_i$ and $b_i$, the objective function will have different hessians for the localized convex functions. Therefore, the search space comprises convex regions in various widths and scales. As a result OHM is a necessary initializer to generate initializing seeds (solutions) in multiple scales to speed up finding the global optima. Obviously, PSO (or ICA) cannot generate seed as easy as OHMPSO, due to its single-scale behavior. It either misses initializing in low-width convex regions, or wastes individual seeds all scattered over wide convex regions. ERP is the average of a distinct number of epochs. This process is to lower the total number of epochs either for the sake of increasing precision or for passing a smaller number of covariance matrices to CSP algorithm. Negative of (11) is passed to the hybrid optimizers to perform minimization.

Using Rayleigh Quotient, both of the following equations when having the weights $a_i$ and $b_i$ can be simplified to a generalized eigenvalue problem and the projectors be optimized completely in a certain way. That makes the only uncertain part of optimization the process of finding $a_i$ and $b_i$ s. Hence, unlike Devlaminck et. al method which has the possibility of getting dumped into local optimal in all its parameters, this form helps the process of optimization be done in a more efficient and robust way while remaining more certain about the result.

**4.1.5 The optimization process of the objective function**
Due to the non-convexity of objective function over weights, the mere usage of GD without powerful random initializer cannot find the global minima and eventually causes the gradient method to undergo premature convergence.

The optimization method alternates between computing CSP using SVD power method with 7 iterations and OHM-GD for updating weights with 6 iterations.

The OHM-GD algorithm is mentioned as below:

---

ALGORITHM 2: The OHM-GD algorithm's pseudo-code

---

Input: $\bar{a}$, $\bar{b}$ : weights in (11) . N : number of GD iterations as 5. F: Objective function in (11) . $\gamma$ : 1e-10 . Parameter settings for OHM.
Output: Optimized $a$, $b$, W

- Zero-Mean and Normalize all mixed components
- According to a number of independent components, randomly draw a positively valued matrix.
- Randomize $a$, $b$ by Nesterov method [26]
- 
- Do until $iter < Max\_Iterations$ or $\sum_{i,j} abs(\bar{a}_{iter} - \bar{a}_{iter-1}) < \gamma$
    - Initialize $a$, $b$ by OHM framework.
    - Do until i < N
        - Compute $\nabla_w F$ using computed gradient.
        - Remove corrupted (high values / NaN values) columns of $\nabla_w F$
        - Depending on type of gradient method, use + or -  $a_{iter} = a_{iter-1} \pm \eta * \nabla_w F$
        - set $b_{iter} = b_{iter-1} \pm \eta * \nabla_w F$
        - Compute new CSP projector using eigenvalue decomposition.
        - set i = i+1

**4.1.6 EEG dataset , assessment method, CSP usage**

EEG data belongs to speech-imagery BCI experiments which are developed by Rostami et.al. [30]. Data is logged using a 16 channeled EEG recorder extracted from 6 subjects, aged between 23 and 30 who performed imagination of vowel sounds. Each subject has taken 180 trials which were approximately 36 trials for the imagination of five class each as a vowel. The sampling rate was 512 Hz for 4 seconds lasted imagery. For evaluating the proposed objective function, the main 5-folded cross-validation process described as follows:

---

ALGORITHM 3: The 5-folded cross-validation pseudo-code

---

- Randomly dissect trials into 5 equal folds of trials.
- Zero-mean and unit variance each channel of data.
- For each selected fold; do:
    - Leave selected fold as the test set and remaining folds as train data.
    - Decimate data by the rate of 8; then bandpass filter data using a fifth order Chebyshev filter with band-pass of [3,30] Hz.
    - Learn a CSP projector using training data folds using CSP algorithm described in 4.1.4.
    - Extract significant components of data out of 16 from both train and test data.
    - Pass results in train and test data to Linear Discriminant Analysis (LDA) or Support Vector Machine (SVM) classifier with linear kernel and LibSVM library [33].
    - Save the validation accuracy
- Average over 5 resulted accuracies.

**4.1.7 Values used for searching and tuning hyper-parameter**

For the GD method, Adam is used due to its adaptive momentum and the weight-decay effect. Weight-decay stops jumping off the local minima by restricting movements towards current direction. The fine-tuning settings the tuner has searched over were {0.2,0.02,0.002} for learning-rate , {0.8,0.9,099} for momentum1 and momentum2 , and also the proposed OHM initializers' reoccurrence occasion are fine-tuned for each 5,10 or 15 GD runs  The best resulted parameters settings are 0.02 for learning-rate, 0.9 for momentum1, 0.8 for momentum2 and 5 for initializer rerun occasion. OHM initializer is tested out in two proposed scenarios OHMPSO and OHMICA with both bases PSO and ICA.

Error-bar in Fig. 4 shows that proposed hybrid optimizer of weighted CSP outperforms other similar averaging approaches like Devlaminck and Lotte (4.1.2 and 4.1.3). All accuracies in the error-bar are validation data averaged over 20 independent runs in first three subjects of speech imagery dataset [30]. All methods except Weighted CSP are evaluated with only single GD without metaheuristic for initialization due to their convex formulation and the fact that all CSP computations have been performed in the same way using SVD power method.

### 4.1.8 Evaluation results and analysis

After designing a suitable objective function for feature learning, the comparison baselines are defined for the OHM. Table 6 shows elapsed time, cost and standard-deviation averaged over 20 independent runs and also best accuracy evaluated on validation data. Two best results per each column are shown in bold. The comparison baselines are single GD [26], Gradient-based PSO (GPSO) [25], Comprehensive Learning PSO (CLPSO) [14] with GD, ICA [12] with GD, CICA [13] with GD and each of which has GD as its main algorithm. GPSO is a combination of GD optimizer with standard PSO which lacks multi-scaled search capability. In GPSO, initialization frequency is controlled by the parameter $N_G$ [25] and it is tuned to 5. CICA, which is a chaotic adjustment of colony direction angle, is selected due to its relevance and performance in chaotic fractal functions and their multi-scale local optimas. Setting PSO/ICA state of the art subsets as the baselines is inappropriate; because due to implementation, runtime, and complexity issues of combining with GD, they are not useful for gradient initializers. Yet, the standard benchmarks' SOTA methods are not necessarily suitable for comparing hybrid-GD approaches; because there are neither multi-modal differentiable benchmark standards to assess existing methods, nor general assessment of hybrid-GD-methods performance in the literature. The selected baselines are most common in machine learning applications, being used as initializers more frequently in the literature [13, 14, 23].

The analysis results show outperformance of both OHMPSO and OHMICA versus other hybrid algorithms especially G-PSO [25]. Also by using standard ICA, CICA, and CLPSO as GD initializer with the same mechanism as of OHMICA-GD, they have failed to outperform our hybrid method.

Due to the formulation of CSP subspace filtering problem, the GD version used in the proposed study was not stochastic like ones used in deep learning; but averaging over sub-batches and saving average cost per batch, may also help big-data models initialization in such case studies.

The validation accuracies also suggest that the decrease in the cost of the objective function over hybrid-GD mode is meaningful and improves the best GD evaluation accuracy mentioned in the 4[th] row of Table 6.

Table 6. Comparison of hybrid optimizers in terms of time, cost, and validation accuracy

| Optimizer method | Avg time | cost mean (Proposed Wgt-CSP objective function. Minus of Obj-fun in Formula 11.) | cost StD. | Best Eval. Accuracy (%) |
|---|---|---|---|---|
| OHMICA-GD* (Proposed) | 0.52 | **-0.90** | 0.05 | **83.24** |
| OHMPSO-GD* (Proposed) | **0.50** | **-0.92** | 0.05 | **86.69** |
| GD | **0.49** | -0.88 | 0.05 | 79.60 |
| CLPSO-GD (CLPSO: [14]) | 0.67 | -0.90 | 0.05 | 81.55 |
| ICA-GD (ICA: [12]) | 0.83 | -0.86 | **0.04** | 75.81 |
| CICA-GD (CICA: [13]) | 1.27 | -0.88 | **0.03** | 81.60 |
| GPSO [25] | 0.59 | -0.89 | **0.04** | 81.70 |

Furthermore, Fig. 4 describes the Comparison of generalization accuracy in our Wgt-CSP method among previous subspace filtering approaches in EEG two-class classification study. Wgt-CSP outperformed other methods mostly when OHMPSO-GD is used as optimizer. Only in one case the result nearly equates normal CSP approach but averaged accuracies of methods described in 4.1.2 and 4.1.3 are less than our objective function. In Fig. 4, standard-deviations in three out of 4 Wgt-CSP modes are relatively low. However, one of them are high and suggests some uncertainties yet have to be tackled by more thorough parameter tuning and preprocessing.

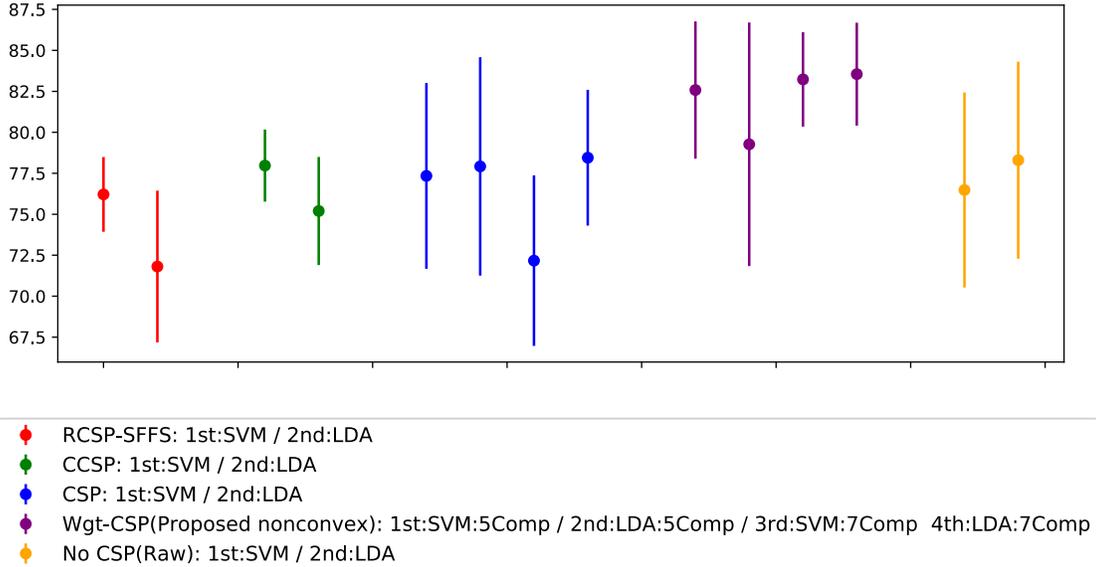

Fig. (4): Comparison of EEG classification accuracy in weighted-CSPs-over-trials (Wgt-CSP) among previous subspace approaches. Wgt-CSP outperformed other methods mostly when used OHMPSO-GD as the optimizer. In the legend, terms 'Comp', 'SVM', 'LDA', 'Raw' mean 'number of projected subspaces', 'SVM classifier', 'Linear Discriminant Analysis', and 'without CSP feature extraction as raw features' respectively.

### 4.2 Comparison of OHM versus ICA/PSO variants on benchmarks in a single setting

The benchmark functions used in this comparison are explained in Tables 7 and 8. Table 7 evaluates single-modal and multi-modal functions while Table 8 lists the stability functions in shifted and rotated modes. P-Values for 20 independent runs over Table 7 and 8 benchmark functions are computed and all p-values are below 0.05 for failing to reject the null-hypothesis of OHM indifference versus other optimizers' results. A set of 20 benchmark function has been used in which their specifications are brought in Tables 4 and 5. Algorithms have been run with NFE (number of function evaluations) destined dependent to corresponding benchmark function. All functions with NFE=30000 had 3-dimensional input vectors. Dimensions for 180000 and 500000 cases were 10 and 30 respectively.

The proposed OHM framework is not fully evaluated on CEC benchmarks over state of the art optimizers for three reasons. First of all, the optimizers baselines in this section are selected based on their usage frequency in gradient-based applications. Secondly, OHMPSO and OHMICA outperform CLPSO and CICA, respectively as elite in CEC competition 2014 [14] and trustworthy in previous neural network case studies [23]. Finally, due to the complexity and time-consuming run-time of state of the art optimizers, the authors refrained to compare the proposed method with them as they are not applicable while being hybridized with GD. The optimizers' baselines used in Table 9 are twofold, ICA and CICA as ICA group, and PSO and CLPSO as PSO group.

The OHMICA and OHMPSO have been compared to proposed improvements of ICA and PSO which are CICA and CLPSO respectively due to more relative simplicity and plausibility for machine learning [13, 14, 23]. Although the OHM is not compared with earlier SOTA methods and they may generally outperform, but they are more computationally intensive, have less literature usage in GD-hybridized settings, and have undergone a more comprehensive hyperparameter tuning. Furthermore, this section also suggests the search space distributions in which the OHM acts more plausibly (multi-modal functions group in this case) and such assessment does not necessarily need SOTA baselines. Analysis results from Table 9 show that multimodal functions Ackley, Rastrigin, Schwefel, and Griewank have overcome these algorithms in OHM in a better way than single-modal counterparts. This verifies the results sought in multi-convex setting of Section 4.1. Its reason is outstanding power of framework to bypass local optimas while its updater chooses a higher level of hierarchy during the search in a sub-hierarchy. This simultaneous searching in levels of hierarchy hinders from getting trapped in local optima.

This analysis is based on algorithms' rank counted per each row of Table 9. Although results show acceptable performance in shifted and rotated functions over robust algorithms like CLPSO, PSO-W, and CICA; but OHMPSO manifested unsatisfactory results in Easom and Quartic. On the other hand, OHMPSO preformed well in multimodal functions like Schwefel. These findings correlate with the aforementioned ideas about cooperation and competition. Cooperative nature of OHMPSO helps it deal better with local search, while competitive nature of OHMICA makes it perform a more promising global search, resulting in better performance in multimodal functions.

Current parameter specifications of the algorithm is unable to handle Beale and Sphere. The more an algorithm acts cautiously for difficult situations, the less it is capable of acting fast in simple functions.

As Table 9 suggests, functions Ackley, Weierestrass, and Griewank have higher relative error improvement in both OHMICA and OHMPSO compared to Rastrigin, Easom, and Schwefel. Weirestrass is a self-similar function that contains the highest density of multi-scale local optimas compared to other benchmarks. Better OHM results in this fractal multi-scale function confirm the multi-scale search capability of our methods.

Moreover, Griewank and Ackley's values are scattered in more scales than Rastrigin, Easom, and Schwefel. This happens because exponential and cosine values are mostly produced by each other in Ackley, Weierestrass, and Griewank functions; while in Rastrigin, Easom, and Schwefel, cosines and exponential functions are summed over each other and that makes high growth rate with multtiple varieties of tones nonexistent in the $2^{nd}$ group . Results suggest there is extra capability in OHM for seeking solutions of fractal multi-scale functions and that once again confirms the necessity for hierarchical searching in nested swarms. This helps the neural nets and highly nonlinear subspace learning search spaces (as in Section 4.1) initialize in a better way.

## Table 7: Benchmark functions to evaluate optimizers, Part 1.

| Function | Range | Desired Optima | Formula | Surface Plot |
|---|---|---|---|---|
| F1) Rosenbrock | $x_i \in [-100,100]$, $i = 1, \ldots, d$ | Min = $\begin{cases} n=2, f(1,1) = 0 \\ n=3, f(1,1,1) = 0 \\ n>3, f(1_{1,1_2}, \ldots, 1_n) = 0 \end{cases}$ | $f_1(x) = \sum_{i=1}^{N-1} 100(x_{i+1} x_i^2)^2 + (x_i)^2$ | |
| F2) Sphere | $x_i \in [-5.12, 5.12]$ | $f(x_1, \ldots, x_n) = f(0, \ldots, 0) = 0$ | $f_2(x) = \sum_{i=1}^{n} x_i^2$ | |
| F3) Dixon Price | $x_i \in [-10,10]$, $i = 1,2$ | $f(x*) = 0$, $X_i = 2^{-\frac{2^i-2}{2^i}}$, $i = 1, \ldots, d$ | $f_3(x) = (x_1 - 1)^2 + \sum_{i=2}^{d} i(2x_i^2 - x_{i-1})^2$ | |
| F4) Beale | $x_i \in [-4.5, 4.5]$, $i = 1,2$ | $f(3, 0.5) = 0$ | $f_4(x, y) = (1.5 - x + xy)^2 + (2.25x + xy^2)^2$ | |
| F5) Easom | $x_i \in [-100,100]$, $i = 1, \ldots, d$ | $f(0, \ldots, 0) = 0$ | $f_5(x) = -\cos(x_1)\cos(x_2)e^{-(x_1-\pi)^2-(x_2-\pi)^2}$ | |
| F6) Quartic | $x_i \in [-1.28, 1.28]$, $i = 1, \ldots, d$ | $f(0, \ldots, 0) = 0$ | $f_{6(X)} = \sum_{i=1}^{n} i x_i^4$ | |
| F7) Schwefel | $x\_i \in [-0.5, 0.5]$, $i = 1, \ldots, d$ | $f(420.9687, \ldots, 420.9687) = 0$ | $f_7(x) = 418.9829d - \sum_{i=1}^{d} x_i \sin\left(\sqrt{|x_i|}\right)$ | |
| F8) Weierstrass | $x_i \in [-0.5, 0.5]$, $i = 1, \ldots, d$ | $f(x_{opt}) = f(o) = x_{bias} = -0.5$ | $f_{8(x)} = \sum_{i=1}^{D} \left( \sum_{k=0}^{k_{max}} a^k \cos(2\pi b^k (x_i + 0.5)) \right) - D \cdot \sum_{k=0}^{k_{max}} a^k \cos(2\pi b^k \cdot 0.5)$ | |
| F9) Rastrigin | $x_i \in [-5.12, 5.12]$, $i = 1, \ldots, d$ | $f(0, \ldots, 0) = 0$ $a = 0.5, b = 3. K_{max} = 20$ | $f_{9(x)} = \sum_{i=1}^{N} (x_i^2 10 \cos(2\pi i x) + 10)$ | |
| F10) Ackley | $x_i \in [-32.768, 32.768]$, $i = 1, \ldots, d$ | $f(0, \ldots, 0) = 0$ $a = 20, b = 0.2$ and $c = 2\pi$ | $f_{10(x)} = -ae^{-b\sqrt{\frac{1}{d} \times \sum_{i=1}^{d} x_i^2}} e^{\frac{1}{d} \times \sum_{i=1}^{d} \cos(cx_i)} + a + e$ | |
| F11) Griewank | $x_i \in [-600, 600]$, $i = 1, \ldots, d$ | $f(0, \ldots, 0) = 0$ | $f_{11(x)} = \sum_{i=1}^{D} \frac{x_i^2}{4000} - \prod_{i=1}^{D} \cos\left(\frac{x_i}{\sqrt{i}}\right)$ | |

**Table 8- Benchmark functions used to assess optimizer's stability. F12 to F16 are rotated functions and F17 to F20 are shifted rotated functions which are both for stability testing purposes. In upcoming results, Types are shown in their own color to be discerned more easily.**

| Function | Range | Desired Optima | Formula | Surface Plot |
|---|---|---|---|---|
| F12) Rotated Ackley | $x_i \in [-32.768, 32.768], i = 1, \ldots, d$ | $f(0, \ldots, 0) = 0$ | $f_{12(x)} = -20.e^{-0.2\sqrt{\frac{1}{D} \times \sum_{i=1}^{D} z_i^2}} e^{\frac{1}{D}\sum_{i=1}^{D} \cos(2\pi z_i)} + 20 + e + f_{opt}$ | 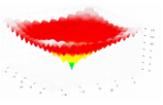 |
| F13) Rotated Rastrigin | $x_i \in [-5.12, 5.12], i = 1, \ldots, d.$ $z = R(0.0512.(x - x_{\_}opt))$ | $f(0, \ldots, 0) = 0$ | $f_{13}(x) = \sum_{i=1}^{D} (z_i^2 10\cos(2\pi z_i) + 10) + f_{opt}$ | 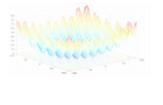 |
| F14) Rotated Schwefel | $x_i \in [-500, 500], i = 1, \ldots, d.$ | $f(x) = 0, x = \frac{1}{6} \times R^{-1} \times [420.9687, \ldots, 420.9687]^T$ | $f_{14(x)} = 418.9829d - \sum_{i=1}^{d} z_i \sin(\sqrt{|z_i|}), z = R(6 \times x)$ | 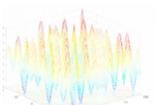 |
| F15) Rotated Griewank | $x_i \in [-600, 600], i = 1, \ldots, d.$ | $f(0, \ldots, 0) = 0$ | $f_{15(x)} = \sum_{i=1}^{D} \frac{z_i^2}{4000} - \prod_{i=1}^{D} \cos\left(\frac{z_i}{\sqrt{i}}\right) + 1 + f_{opt}, z = R(6.x)$ | 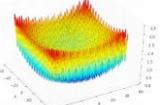 |
| F16) Rotated Weierstrass | $x_i \in [-0.5, 0.5]$ $a = 0.5, b = 3. K_{\_}max = 20, z = R(0.005.x)$ | $f(0, \ldots, 0) = 0, sourcefunction$ | $\sum_{i=1}^{D} \left(\sum_{k=0}^{k_{max}} a^k \cos(2\pi b^{k(z_i+0.5)})\right) - D.\sum_{k=0}^{k_{max}} a^k \cos(2\pi b^k . 0.5) + f_{opt}$ | 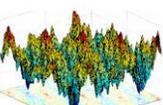 |
| F17) Rotate Shift Expand Scaffer | $x_i \in [-100, 100]$ | $f(0, \ldots, 0) = 0, sourcefution$ | $\sum_{i=1}^{D-1} p(z_i, z_{i+1}) + p(z_D, z_1)$ $p(u, y) = \frac{\sin(\sqrt{u^2+y^2}) - 0.5}{(1 + 0.001.(u^2+y^2))^2} + 0.5$ | 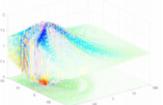 |
| F18) Rotate Shift Griewank | $x_i \in [-600, 600], i = 1, \ldots, d.$ | $f(0, \ldots, 0) = 0, sourcefunction$ | $\sum_{i=1}^{D} \frac{z_i^2}{4000} - \prod_{i=1}^{D} \cos\left(\frac{z_i}{\sqrt{i}}\right)$ $z = R(6.(x - x_{\_}opt))$ | 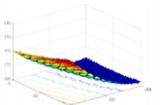 |
| F19) Rotate Shift Rastrigrin | $x_i \in [-5.12, 5.12], i = 1, \ldots, d.$ | $f(0, \ldots, 0) = 0, sourcefunction$ | $f_{19(x)} = \sum_{i=1}^{D} (z_i^2 10\cos(2\pi z_i) + 10) + f_{(bias)_{10}}$ | 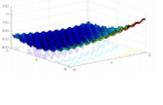 |
| F20) Rotate Shift Ackly | $x_{\_}i \in [-32.768, 32.768], i = 1, \ldots, d$ | $f(0, \ldots, 0) = 0, sourcefunction$ | $f_{20(x)} = -20e^{-0.2\sqrt{\frac{1}{D} \times \sum_{i=1}^{D} z_i^2}} e^{\frac{1}{d} \times \sum_{i=1}^{d} \cos(2\pi z_i)} + 20 + e + f_{\_}(bias)$ | 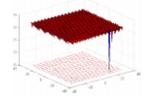 |

**Table 9: Optimization results. Comparisons of error from global minima in proposed OHMPSO and OHMICA algorithms versus known optimizer, those of more similar category. All functions with NFE=30000 had 3 dimensional input vectors. Dimensions for 180000 and 500000 cases were 10 and 30 respectively.**

|  | NFE | PSO | | CLPSO | | ICA | | CICA | | OHM-ICA* | | OHM-PSO* |
|---|---|---|---|---|---|---|---|---|---|---|---|---|
| Rosenbrock | 180000 | 2.14E+00<br>1.81E+00 | ± | 2.46E+00<br>1.70E+00 | ± | 2.00E-01<br>3.60E-01 | ± | 2.00E-02<br>2.00E-02 | ± | 6.15E-01<br>1.35E-04 | ± | **2.03E-11**<br>**2.35E-12** |
| Sphere | 180000 | N/A<br>N/A | ± | **5.15E-29**<br>**2.16E-28** | ± | 2.87E+01<br>2.07E+00 | ± | 2.50E-07<br>1.10E-01 | ± | 5.87E-05<br>1.23E-06 | ± | 9.94E-03<br>7.48E-04 |
| Dixon Price | 500000 | 1.00E+00<br>5.05E-02 | ± | 1.32E-05<br>6.95E-07 | ± | 9.11E+00<br>3.73E-01 | ± | 6.28E-02<br>5.07E-03 | ± | **1.58E-03**<br>**3.99E-05** | ± | **1.00E-09**<br>**9.04E-11** |
| Beale | 500000 | 1.42E+01<br>1.28E+00 | ± | 9.04E-02<br>9.79E-03 | ± | 5.05E+02<br>2.53E+00 | ± | 9.40E-01<br>6.32E-02 | ± | **9.29E-05**<br>**1.92E-06** | ± | **2.39E-06**<br>**3.03E-07** |
| Easom | 500000 | 2.68E-09<br>3.34E-10 | ± | 6.52E-08<br>9.39E-09 | ± | 6.26E-11<br>7.71E-12 | ± | 3.27E-06<br>4.29E-07 | ± | **1.87E-09**<br>**1.81E-13** | ± | **6.68E-12**<br>**8.21E-13** |
| Quartic | 500000 | N/A<br>N/A | ± | N/A | ± | N/A<br>N/A | ± | N/A<br>N/A | ± | N/A<br>N/A | ± | **1.14E-13**<br>**6.85E-14** |
| Schwefel | 180000 | 9.82E-02<br>7.21E-03 | ± | N/A | ± | 5.99E-01<br>6.66E-02 | ± | 1.98E+04<br>7.94E+01 | ± | **1.10E-12**<br>**5.21E-11** | ± | **3.51E-15**<br>**2.76E-16** |
| Weierstrass | 180000 | N/A<br>N/A | ± | N/A | ± | 5.51E-02<br>2.09E-01 | ± | 1.29E-04<br>6.60E-04 | ± | **3.26E-07**<br>**3.79E-05** | ± | **3.20E-11**<br>**3.37E-13** |
| Rastrigin | 500000 | 9.95E-01<br>6.09E-01 | ± | N/A | ± | 1.66E-06<br>9.12E-06 | ± | 9.34E-09<br>3.42E-08 | ± | **5.19E-15**<br>**9.25E-16** | ± | **2.37E-14**<br>**1.19E-15** |
| Ackley | 500000 | N/A<br>N/A | ± | 4.32E-14<br>2.55E-14 | ± | 7.11E-05<br>8.20E-06 | ± | 1.02E-07<br>1.23E-07 | ± | **1.29E-18**<br>**1.19E-19** | ± | **2.64E-19**<br>**3.72E-19** |
| Griewank | 500000 | 1.14E-01<br>4.96E-02 | ± | 4.56E-03<br>4.81E-03 | ± | 1.03E-10<br>8.14E-10 | ± | 3.47E-14<br>5.07E-15 | ± | **3.12E-18**<br>**1.06E-19** | ± | **1.13E-21**<br>**1.60E-12** |
| Rotated Ackley | 30000 | 1.73E+01<br>3.43E-01 | ± | **3.56E-05**<br>**4.35E-06** | ± | 3.39E+02<br>1.96E+00 | ± | 1.92E+00<br>2.41E-01 | ± | 3.03E-02<br>5.83E-04 | ± | **2.04E-12**<br>**1.86E-13** |
| Rotated Rastrigin | 30000 | 1.62E+01<br>5.45E-01 | ± | 5.97E+00<br>2.93E-01 | ± | 2.48E+02<br>4.94E+00 | ± | 7.51E+01<br>6.17E+00 | ± | **1.41E-01**<br>**3.53E+02** | ± | **8.87E-09**<br>**4.57E-10** |
| Rotated Schwefel | 30000 | 5.35E+03<br>8.45E+01 | ± | **1.14E+02**<br>**1.30E+00** | ± | 2.32E+06<br>8.09E+02 | ± | 1.07E+04<br>1.46E+01 | ± | 3.90E+05<br>7.43E-01 | ± | **3.21E-10**<br>**3.02E-11** |
| Rotated Griewank | 180000 | 1.44E-02<br>8.86E-04 | ± | **4.50E-02**<br>**4.50E-03** | ± | 2.56E-02<br>1.79E-03 | ± | 2.68E+01<br>6.97E-01 | ± | 1.26E+00<br>6.38E-02 | ± | **1.43E-11**<br>**1.76E-12** |
| Rotated Weierstrass | 180000 | 9.32E-06<br>7.58E-07 | ± | **3.72E-10**<br>**5.23E-11** | ± | 6.83E-07<br>3.55E-08 | ± | 6.04E+01<br>4.14E+00 | ± | 7.33E-09<br>7.59E-06 | ± | **4.69E-17**<br>**6.09E-18** |
| Rotate Shift ExpandScaffer | 30000 | 4.90E+00<br>1.72E-01 | ± | 1.76E+00<br>1.65E-01 | ± | 4.82E+01<br>2.18E+00 | ± | 4.12E+01<br>6.27E-01 | ± | **1.47E-01**<br>**8.38E-04** | ± | **4.89E-13**<br>**5.52E-14** |
| Rotate Shift Griewank | 30000 | 5.38E+01<br>6.59E+00 | ± | 3.55E+01<br>7.02E-01 | ± | 2.74E+03<br>3.80E+01 | ± | 6.52E+02<br>3.74E+00 | ± | **5.35E+00**<br>**7.41E-02** | ± | **6.02E-11**<br>**6.69E-12** |
| Rotate Shift Rastrigrin | 30000 | 6.92E+02<br>2.57E+01 | ± | **6.01E+01**<br>**1.78E+00** | ± | 2.66E+04<br>2.37E+01 | ± | 2.41E+04<br>1.81E+02 | ± | 1.84E+03<br>2.85E+02 | ± | **1.82E-11**<br>**2.54E-12** |
| Rotate Shift Ackly | 180000 | 2.07E+01<br>2.76E+00 | ± | **7.20E+00**<br>**4.15E-01** | ± | 2.81E+02<br>2.92E+00 | ± | 1.24E+02<br>2.11E+00 | ± | 2.54E+04<br>1.84E-02 | ± | **2.10E-12**<br>**2.50E-13** |

## 4.3 Comparison between OHM variants over selected benchmark functions in single mode

As mentioned in the beginning of Section 4, in self-tuning-mode of OHM, hierarchy level selector uses fitness distribution of currently sought solutions by a Roulette Wheel Selector to select hierarchy level. When this mode is active, the algorithm adaptively changes the significance of each hierarchy level. Performance comparison out of Table 10 shows outperformance of self-tuning case of OHM framework with normal OHM. In both cases of ICA based hierarchy and PSO based hierarchy, the function "Beale" is seen to have its performance improved. Beale function is multimodal with sharp peaks which faster finding of its global optima has been achieved through automatic adaptation of the significance of hierarchy levels.

The self-tuning process worked fairly efficient for OHMPSO and led to negative results for the case of OHMICA. One reason could be that constant changing of hierarchy level effectiveness misguide competition process of OHMICA and combines and finally fades out borders among organizations. Another reason could be because of inappropriate initialization of level effectiveness. Because the initialization parameter for that was not tuned for the best contrarily to other parameters.

## 5 Conclusion

This paper introduces a new metaheuristic framework for basic PSO and ICA. The main objective is to improve GD initialization power using nested sub-hierarchies and super-hierarchies search used by PSO and ICA generalizations respectively. This framework has many applications in multi-convex and gradient-based models. Proposed algorithms out of framework have been tested out on a hybrid subspace learning study and also a single study with 20 benchmark functions over optimizers in a similar context. The results are satisfactory in improving GD, GPSO, and other hybrid algorithms. Insights from a single mode of benchmark evaluations unravel OHMICA's performance for multi-modal cases while suggests that OHMPSO acts relatively more acceptable in stability and also exploration with exploitation.

In upcoming works, a more thorough parameter tuning will be approached for OHM to compare it with SOTA methods by CEC competition benchmarks for non-hybrid case studies used for tuning. OHM will be used in batched large-scale data in deep learning scenarios which need random initializer for Stochastic GD.

**Table 10: Comparison of variations of proposed framework introduced in the table. The N/A values returned error during runtime.**

|  | NFE | OHMPSO* | OHMPSO-ST | OHMICA* | OHMICA-ST |
|---|---|---|---|---|---|
| Rosenbrock | 180000 | 2.03E-11 ± 2.35E-12 | **2.03E-06±1.93E-07** | 6.15E-01±1.35E-04 | 8.30E-03±3.76E-06 |
| phere | 180000 | 9.94E-03±7.48E-04 | 9.94E-04±1.30E-05 | 5.87E-05±1.23E-06 | **5.04E-06±9.40E-08** |
| Dixon Price | 500000 | **1.00E-09±9.04E-11** | 1.00E-05±1.18E-06 | 1.58E-03±3.99E-05 | 6.90E-05±7.39E-03 |
| Beale | 500000 | **2.39E-06±3.03E-07** | 2.39E-13±1.49E-14 | 9.29E-05±1.92E-06 | 1.47E+00±2.74E-01 |
| Easom | 500000 | **6.68E-12±8.21E-13** | 6.68E-10±5.95E-11 | 1.87E-09±1.81E-13 | 1.53E-10±8.71E-12 |
| Quartic | 500000 | 1.14E-13±6.85E-14 | **1.14E-42±1.68E-43** | N/A±N/A | N/A±N/A |
| Schwefel | 180000 | **3.51E-15±2.76E-16** | 3.51E-10±2.09E-11 | 1.10E-12±5.21E-11 | 2.11E+00±4.01E-05 |
| Weierstrass | 180000 | **3.20E-11±3.37E-13** | 3.20E-05±1.80E-06 | 3.26E-07±3.79E-05 | 1.47E-08±2.48E-03 |
| Rastrigin | 500000 | 2.37E-14±1.19E-15 | 2.37E-14±3.36E-10 | 5.19E-15±9.25E-16 | **1.26E-15±6.25E-16** |
| Ackley | 500000 | **2.64E-19±3.72E-19** | 2.64E-19±2.78E-11 | 1.29E-18± 1.19E-19 | 1.14E-18±7.79E-17 |
| Griewank | 500000 | **1.13E-21±1.60E-12** | 1.13E-18±6.17E-09 | 3.12E-18±1.06E-19 | 2.83E-16±1.37E-18 |
| Rotated Ackley | 30000 | **2.04E-12±1.86E-13** | 2.04E-07±1.91E-08 | 3.03E-02±5.83E-04 | 9.53E-04±5.52E-05 |
| Rotated Rastrigin | 30000 | **8.87E-09±4.57E-10** | 8.87E-06±9.68E-07 | 1.41E-01±3.53E+02 | 5.26E-04±2.08E+00 |
| Rotated Schwefel | 30000 | **3.21E-10±3.02E-11** | 3.21E-08±2.24E-09 | 3.90E+05±7.43E-01 | 1.63E+04±1.48E-02 |
| Rotated Griewank | 180000 | **1.43E-11±1.76E-12** | 1.43E-08±1.04E-09 | 1.26E+00±6.38E-02 | 1.70E-01±8.57E+00 |
| Rotated Weierstrass | 180000 | **4.69E-17±6.09E-18** | 4.69E-13±4.01E-14 | 7.33E-09±7.59E-06 | 1.24E-06±5.07E-08 |
| Rot Shft Ex Scaffer | 30000 | **4.89E-13±5.52E-14** | 4.89E-08±2.91E-09 | 1.47E-01±8.38E-04 | 4.94E-04±3.89E-05 |
| Rotate Shift Griewank | 30000 | **6.02E-11±6.69E-12** | 6.02E-09±4.12E-10 | 5.35E+00±7.41E-02 | 5.92E-02±9.62E-04 |
| Rotate Shift Rastrigrin | 30000 | **1.82E-11±2.54E-12** | 1.82E-09±9.10E-11 | 1.84E+03±2.85E+02 | 3.82E+01±8.47E+00 |
| Rotate Shift Ackly | 180000 | **2.10E-12± 2.50E-13** | 2.10E-05 ±2.80E-06 | 2.54E+04±1.84E-02 | 8.28E+02±1.48E+00 |